\RequirePackage{fix-cm}
\documentclass[smallextended]{svjour3}

\smartqed  

\usepackage{graphicx}
\usepackage{amsmath}
\usepackage{amssymb}
\usepackage{algorithmic}
\usepackage{algorithm}
\usepackage{subfigure}
\usepackage{array}
\usepackage{tabularx}
\usepackage{multirow}
\usepackage{pdfsync}
\usepackage[pagewise]{lineno}
\usepackage{color}

\newcommand{\bfx}{{\textbf{x}}}
\newcommand{\bfv}{{\textbf{v}}}

\newcommand{\bfw}{{\textbf{w}}}

\newcommand{\bfz}{{\textbf{z}}}

\newcommand{\bfphi}{{\boldsymbol{\phi}}}

\newcommand{\bfpsi}{{\boldsymbol{\psi}}}

\begin{document}

\title{Domain Transfer Multi-Instance Dictionary Learning}

\author{Ke Wang \and Jiayong Liu \and Daniel Gonz{\'a}lez}

\institute{
Ke Wang\at
College of Mathematics, Sichuan University, Chengdu 610064, China\\
\and
Jiayong Liu$*$\at
College of Electronics and Information Engineering, Sichuan University, Chengdu 610064, China\\
\email{jiayongliuscu@gmail.com}\\
$^*$ Corresponding author
\and
Daniel Gonz{\'a}lez\at
Computer Science Department, Catholic University of Murcia, Murcia 30107, Spain\\
\email{danielgonzalezmurcia@gmail.com}
}

\date{Received: date / Accepted: date}

\maketitle

\begin{abstract}
In this paper, we invest the domain transfer learning problem with multi-instance data. We assume we already have a well-trained multi-instance dictionary and its corresponding classifier from the source domain, which can be used to represent and classify the bags. But it cannot be directly used to the target domain. Thus we propose to adapt them to the target domain by adding an adaptive term to the source domain classifier. The adaptive function is a linear function based a domain transfer multi-instance dictionary. Given a target domain bag, we first map it to a bag-level feature space using the domain transfer dictionary, and then apply a the linear adaptive function to its bag-level feature vector. To learn the domain-transfer dictionary and the adaptive function parameter, we simultaneously minimize the average classification error of the target domain classifier over the target domain training set, and the complexities of both the adaptive function parameter and the domain transfer dictionary. The minimization problem is solved by an iterative algorithm which update the dictionary and the function parameter alternately. Experiments over several benchmark data sets show the advantage of the proposed method over existing state-of-the-art domain transfer multi-instance learning methods.
\end{abstract}


\section{Introduction}

Machine learning usually assumes that the training data are from the same domain \cite{Redko201635,fan2011margin,fan2015improved,fan2010enhanced,fan2014finding,fan2014tightening,wang2014effective,wang2015supervised,liu2015supervised,7379459,lin2016multi}.
Recently, transfer learning has been a well-studied topic in machine learning community. It refers to the problem of learning a machine learning modeling for a target domain with help from a source domain \cite{wang2014domain,Lu201683,Mei2015,Yang20154701,Zhao201660}. For a machine learning problem, a source domain is a domain with sufficient training data, which are well labeled, and thus it is easy to train models in this domain. A target domain is usually in lack of training data, or in lack of labels. Because of the lack of data and/or label, the training in the target domain is difficult, and has the problem of over-fitting. The source domain and target domains share the same feature space and label space, thus it would be very helpful to use the source domain data to help the learning problem in the target domain. However, the data distributions of the source and target domains are usually very different, if we simply use the models trained by using the source domain training set, the prediction performance over the target domain can be very inferior. The reason for this is that the difference between the two different domain distributions can even be more significant than the difference between different classes. To solve this problem, we need to transfer the source domain model to the target domain by adapting the source domain predictor to the target domain data set. This is called domain transfer learning. Given a source domain predictor, its train a target domain predictor by adapting it to the target domain over the target domain training set, even when the target domain training set itself is insufficient to train a new target domain predictor. One example of domain transfer learning is spoken Arabic digit recognition \cite{hammami2010improved,hammami2009tree}. In a spoken Arabic digit data set, there are two domains of data, which are male voice signal, and female voice signal. For each domain, there are voice data of ten classes, and each class is a one digit, varying from one to ten. Although the voice signal of both domains shares the same feature space and class label space, however, the classifier trained with male domain data cannot be used directly in the female voice signal domain, due to the significant difference between male and female voice signal domains. Thus an adaptation is necessary to transfer the classifier for the male domain classifier to the female domain. Another example of transfer learning is spam detection problem \cite{duan2012domain}. For one email user, we can train a spam detector from the well labeled spam and normal emails. However, it cannot be used to detect the spams of another user because the emails of these two users may are significantly different from each other. Thus to use the first user's detector to the second user's email, we need to perform an adaptation to the detector.

Up to now, most of the domain transfer learning methods are focused on single instance data \cite{duan2009domain,duan2012domain,ling2008spectral,yang2007cross,pan2010cross}. That is, one data point is only presented as one single instance. However, in many machine learning applications, one data point can contains multiple instances, and is presented as a bag of instances. This type of data is called multi-instance data. In domain transfer learning problem, many applications require to learn domain transfer classifier for multi-instance data. However, only a few works are done to this direction, while most existing works ignore the nature of multi-instance data and only treat them as one single feature vector. To overcome this problem, two methods have been proposed for domain transfer learning with multi-instance data.

\begin{itemize}
\item Zhang and Si \cite{zhang2009multiple} proposed the problem of domain transfer learning for multi-instance data, and  a novel method to solve it within the framework of multi-task learning. More than one source domains are considered and each domain is considered as a task. The target domain classifier is obtained by a weighted linear combination of the classifiers of the multiple tasks.

\item Wang et al. \cite{wang2014adaptive} proposed an adaptive knowledge transfer learning framework for multi-instance data. This method adapts the source-target domain cross class knowledge to the target domain for the multi-instance data to boost the learning. It also build a data-dependent mixture model to combine the knowledge from the source domain to enhance the target domain weak classifier. The objective of this model is optimized by an iterative coordinate descent method as a constraint concave-convex programming problem.
\end{itemize}

It has been shown that the most effective way for multi-instance learning method is using multi-instance dictionary to map a bag of instances to a bag-level feature vector \cite{chen2006miles,fu2011milis}. This representation method is based on bag-instance similarity, and the critical component of this method is the learning of the dictionary. In this paper, we invest the problem of domain transfer multi-instance learning problem and propose a novel solution based on the dictionary. According to our knowledge, this is the first work toward the direction of learning multi-instance dictionary in the scene of domain transfer learning.

We assume that in the source domain, we have a well-trained multi-instance dictionary and a corresponding classifier. Using this dictionary, we can map a bag of instances to a bag-level feature by using the bag-instance similarity, and then in the bag-level feature space, we can apply the classifier to classify the bag. However, the source domain dictionary and classifier cannot be directly applied to the target domain, we propose to add an adaptive term to the source domain classifier to construct the target domain classifier to transfer the knowledge of the source domain to the target domain. To construct the domain adaptive term, we propose to learn a domain transfer dictionary to represent a target domain bag to the bag-level feature space, and the adaptive term is designed to be a linear function in this space. The problem is transferred to the problem of learning the domain transfer dictionary and adaptive function parameter. To this end, we propose to optimize both the mentioned parameters by learn from a limited target domain training set. We propose to minimize the hinge loss of the target domain training bags, the complexity of the adaptive function, and the complexity of the domain transfer dictionary. To solve the minimization problem, we develop an iterative algorithm. The contributions of this paper are listed as follows.

\begin{itemize}
\item \textbf{Contribution \#1}. We propose a novel problem of transfer domain dictionary learning. This problem is beyond simple domain transfer learning or dictionary learning. It is proposed to explore the cross-domain knowledge in the aspect of dictionary learning. Its significance is to learn the critical dictionary which can bridge the source and target domains. It only learns a good predictor in the target domain with help of source domain, but also reveals the nature of the multi-instance data which connects the to domains.
\item \textbf{Contribution \#2}. We build a novel learning model to solve this problem. This model based on a joint optimization problem of both the domain transfer dictionary and the domain adaptation function. The objective is composed of three terms. The first term is the average classification error over the target domain training bags, measured by the hinge loss, the second term is the complexity term of the adaptation function, and the last term is the complexity term of the domain transfer dictionary. In this way, our problem is modeled as a minimization problem.

\item \textbf{Contribution \#3}. We also develop an effective algorithm to solve the problem proposed in \textbf{Contribution \#2}. We use the Lagrange multiplier method and an alternate optimization strategy. The optimization of the adaptation function parameter is transferred to the optimization of some Lagrange multipliers, and the optimization of domain transfer dictionary and the Lagrange multipliers are conducted alternately. The optimization of Lagrange multipliers is performed as a quadratic programming problem, while the domain transfer dictionary is updated by gradient descent algorithm.
\end{itemize}

This paper is organized in the following way. In section \ref{sec:method}, we introduce the proposed domain transfer dictionary learning method, in section \ref{sec:experiment}, we give the experimental results of the proposed method over several benchmark data sets, and finally in section \ref{sec:conclusion}, we conclude the paper with some future works.

\section{Learning Domain Transfer Dictionary}
\label{sec:method}

In this section, we introduce the newly proposed multi-instance dictionary learning method for domain transfer learning problem. In section \ref{sec:objective}, we model the learning problem as a minimization problem, and the problem is solved in section \ref{sec:optimization}. Furthermore, in section \ref{sec:algorithm}, we propose an iterative algorithm to implement the solution of the problem.

\subsection{Problem modeling}
\label{sec:objective}

In the problem of domain transfer multi-instance learning, we suppose we have two domains, which are a source domain and a target domain. For the source domain, we have a training set, and we have learned a well-trained multi-instance dictionary, $\Phi= \{\bfphi_k\}_{k=1}^{\iota}$, where $\bfphi_k\in \mathbb{R}^d$ is the vector of the $k$-th word of the dictionary, and $\iota$ is the number of the words in this dictionary. With this dictionary, we can represent a bag of instances as a feature vector of $\iota$ dimensions. Suppose we have a bag denoted as $B=\{\bfx_j\}_{j=1}^m$, where $\bfx_j$ is $d$-dimensional feature vector the $j$-th instance of the bag, and $m$ is the number of the instances of this bag, we can represent as

\begin{equation}
\begin{aligned}
B\rightarrow \bfz_B^\Phi = \left[\max_{j=1}^m \bfphi_1^\top\bfx_j, \cdots, \max_{j=1}^m \bfphi_\iota^\top\bfx_j\right]^\top \in \mathbb{R}^\iota
\end{aligned}
\end{equation}
where the $k$-th dimension is the maximum dot-produce between the $k$-th codeword and the instances of $B$. {The motive to use the maximum dot-product to measure the similarly between an instance and a bag is it is simple and parameter-free. Some other similarity measures such as Gaussian kernel requires additional parameters such as the band-width parameter. The tuning of these parameters is time-consuming and has the problem of over-fitting. The chosen similarity maximum dot-product does not have such problems because it is parameter-free. }With this bag-level feature vector, we can classify the bags in the bag-level space. To this end, we also have a well-trained classifier, $f(B)$, which is learned from the source domain, to map the bag $B$ to its true binary class label, $y\in \{+1,-1\}$, from its bag-level features,

\begin{equation}
\begin{aligned}
y\leftarrow f(B; \Phi, \bfv) = \bfv^\top \bfz_B^\Phi,
\end{aligned}
\end{equation}
where $\bfv\in \mathbb{R}^\iota$ is the parameter of the source domain classifier. The source domain dictionary, $\Phi$, and its corresponding bag-level classifier, $\bfv$ are trained over the source domain, and are supposed work well in the source domain. Moreover, we have a target domain, and we also want to present and classify the bags of the target domain. One direct method is to apply the source domain dictionary and classifier to the target domain data. However, this is not suitable due to the significant difference between the distributions of the source domain and target domain. We propose to design the target domain multi-instance classifier, $g$, by adapting the source domain dictionary classifier, $f$, to the target domain. More specifically, the target domain classifier is the combination of the source domain classifier and an adaptation term, $\Delta$,

\begin{equation}
\label{equ:g}
\begin{aligned}
g(B; \Psi, \bfw) = f(B; \Phi, \bfv) + \Delta(B; \Psi, \bfw)
\end{aligned}
\end{equation}
where $\Psi$ and $\bfw$ are parameters of the adaptation term which will be specified as follows. $\Psi$ is the domain transfer multi-instance dictionary, and it contains $\kappa$ codewords, $\Psi = \{\bfpsi_k\}_{k=1}^\kappa$, where $\bfpsi_k\in \mathbb{R}^d$ is its $k$-th codeword. We use it to represent a bag $B$ to a $\kappa$-dimensional bag-level feature vector,

\begin{equation}
\begin{aligned}
B\rightarrow \bfz_B^\Psi = \left[\max_{j=1}^m \bfpsi_1^\top\bfx_j, \cdots, \max_{j=1}^m \bfpsi_\kappa^\top\bfx_j\right]^\top \in \mathbb{R}^\kappa.
\end{aligned}
\end{equation}
The adaptation term is a linear function in this bag-level feature space constructed by the domain transfer dictionary, $\Psi$,

\begin{equation}
\label{equ:delta}
\begin{aligned}
\Delta(B; \Psi, \bfw)
&= \bfw^\top \bfz_B^\Psi
\end{aligned}
\end{equation}
$\bfw = [w_1, \cdots, w_\kappa] \in \mathbb{R}^\kappa$ is the $\kappa$-dimensional adaptation function parameter vector, and $w_k$ is its $k$-th element. Substituting the definition of the adaptation term $\Delta$ to (\ref{equ:g}), we can rewrite the target domain classifier $g$ as follows,

\begin{equation}
\label{equ:g}
\begin{aligned}
g(B; \Psi, \bfw) = f(B; \Phi, \bfv) + \bfw^\top \bfz_B^\Psi,
\end{aligned}
\end{equation}
and the problem of domain transfer multi-instance dictionary learning is to learn both $\Psi$ and $\bfw$ by using a target domain training set. The target domain training set is denoted as $X=\{(B_i,y_i)\}_{i=1}^n$, where $B_i$ is the bag of the $i$-th target domain training data point, and $y_i\in \{+1,-1\}$ is the binary class label of the $i$-th target domain training data point. $B_i$ contains $m_i$ instances, $B_i = \{\bfx^i_j\}_{j=1}^{m_i}$, where $\bfx^i_j\in \mathbb{R}^d$ is the $j$-th instance of the $i$-th bag. Given a dictionary $\Psi$, without confusion, the bag-level vector of $B_i$, $\bfz_{B_i}^\Psi$ is simply denoted as $\bfz_i^\Psi$. Moreover, the classification response of the source domain classifier over this bag is denoted as

\begin{equation}
\begin{aligned}
f_i = f(B_i; \Phi, \bfv) = \bfv^\top \bfz_i^\Phi.
\end{aligned}
\end{equation}
In this way, the response of the target domain classifier over $B_i$ is given as

\begin{equation}
\label{equ:g}
\begin{aligned}
g(B_i; \Psi, \bfw) = f_i + \bfw^\top \bfz_i^\Psi,
\end{aligned}
\end{equation}

To learn both the domain transfer dictionary $\Psi$ and the classifier parameter $\bfw$ over the target domain training set, we consider the following three minimization problems.

\begin{itemize}
\item \textbf{Minimization of the classification errors}. The classification error of the $i$-th training bag is measure by the hinge loss,

\begin{equation}
\label{equ:hinge}
\begin{aligned}
E(g(B_i; \Psi, \bfw), y_i)
=& \max \left ( 0, 1 - y_i g(B_i; \Psi, \bfw)  \right )\\
=& \max \left ( 0, 1 - y_i \left ( f_i + \bfw^\top \bfz_i^\Psi \right ) \right )
\end{aligned}
\end{equation}
To seek the optimal classifier and dictionary, this classification error term should be minimized. We propose to minimize the average hinge loss over the target domain training set,

\begin{equation}
\label{equ:hinge1}
\begin{aligned}
\min_{\Psi,\bfw} \left \{ \frac{1}{n} \sum_{i=1}^n E(g(B_i; \Psi, \bfw), y_i)
= \frac{1}{n} \sum_{i=1}^n \max \left ( 0,  1 - y_i \left ( f_i + \bfw^\top \bfz_i^\Psi \right )  \right )\right \}.
\end{aligned}
\end{equation}
The minimization of this problem is difficult because it is couple with a maximization problem within the hinge loss function. To solve this problem, we introduce a slack variable $\xi_i$ to present the maximum variable between $0$ and $1 - y_i \left ( f_i + \bfw^\top \bfz_i^\Psi \right )$,

\begin{equation}
\label{equ:xi}
\begin{aligned}
\xi_i \geq 0, \xi_i \geq 1 - y_i \left ( f_i + \bfw^\top \bfz_i^\Psi \right ).
\end{aligned}
\end{equation}
With this slack variable, we can rewrite the problem in (\ref{equ:hinge1}) as a constrained minimization problem as follows,

\begin{equation}
\label{equ:hinge2}
\begin{aligned}
\min_{\Psi,\bfw,\xi_i|_{i=1}^n}&~ \frac{1}{n} \sum_{i=1}^n \xi_i,\\
s.t.&~ \xi_i \geq 0, \xi_i \geq 1 - y_i \left ( f_i + \bfw^\top \bfz_i^\Psi \right ), \forall ~i=1,\cdots,n.
\end{aligned}
\end{equation}

\item \textbf{Reducing the complexity of the adaptation function}. To prevent the over-fitting problem, we want to keep the adaptation function as simply as possible, and reduce the complexity of the adaptation function. The complexity of the adaptation function is measured by the squared $\ell_2$ norm of the adaptation function parameter,

\begin{equation}
\label{equ:w}
\begin{aligned}
R(\bfw) = \frac{1}{2} \|\bfw\|_2^2 = \frac{1}{2} \sum_{k=1}^\kappa w^2_k.
\end{aligned}
\end{equation}
To this end, we propose to minimize this regularization term as follows,

\begin{equation}
\label{equ:w1}
\begin{aligned}
\min_{\bfw} \left \{ R(\bfw) = \frac{1}{2} \|\bfw\|_2^2 \right \}
\end{aligned}
\end{equation}
The solution of this single problem is an all-zero vector $\bfw = [0,\cdots,0]$. This solution is not optimal for the overall problem, but when this minimization problem is combined with the other problems, this term can bring a tradeoff between the classification error and the adaptation function simplicity.

\item \textbf{Reducing the complexity of the dictionary}. We also hope the dictionary can be as simple as possible. The complexity is also measure by the squared $\ell_2$ norms of the codewords,

\begin{equation}
\label{equ:Psi}
\begin{aligned}
\min_{\Psi} \left \{
Q(\Psi) = \frac{1}{2} \sum_{k=1}^\kappa \|\bfpsi_k\|_2^2 \right \}
\end{aligned}
\end{equation}
Similarly, the solution for this single problem is $\kappa$ all-zero codewords, $\bfpsi_k = [0,\cdots,0], k=1,\cdots, \kappa$. So we also use it as a regularization term to tradeoff the learning of the dictionary.

\end{itemize}

The overall minimization problem is the weighted linear combination of the problems in (\ref{equ:hinge2}), (\ref{equ:w1}), and (\ref{equ:Psi}).

\begin{equation}
\label{equ:hinge2}
\begin{aligned}
\min_{\Psi,\bfw,\xi_{i=1}^n}~& \left \{ \frac{1}{n} \sum_{i=1}^n \xi_i + C_1 R(\bfw) + C_2 Q(\Psi)\right .\\
&= \left. \frac{1}{n} \sum_{i=1}^n \xi_i + \frac{C_1}{2} \|\bfw\|_2^2  +  \frac{C_2}{2} \sum_{k=1}^\kappa \|\bfpsi_k\|_2^2 \right \}\\
s.t.~& \xi_i \geq 0, \xi_i \geq 1 - y_i \left ( f_i + \bfw^\top \bfz_i^\Psi \right ), \forall ~i=1,\cdots,n,
\end{aligned}
\end{equation}
where $C_1$ is the weight of the regularization term of $\bfw$, and $C_2$ is the weight of the regularization term of $\Psi$. Please note that in the objective function, the first term is the classification error term of the target domain, which is critical for the learning of optimal domain transfer multi-instance dictionary. This term grantees that the learned dictionary and its corresponding classifier can lead to a good classification accuracy over the training set of the target domain. The second and third terms are regularization terms which try to generalize the learned dictionary and classifier to the data beyond the training set.

\subsection{Problem solving}
\label{sec:optimization}

In this section, we discuss how to solve the overall problem in (\ref{equ:hinge2}). We use the Lagrange multiplier method to solve this problem. We first define a Lagrange multiplier variable $\alpha_i\geq 0$ for each constraint $\xi_i \geq 0$, and a Lagrange multiplier variable $\beta_i\geq 0$ for each constraint $\xi_i \geq 1 - y_i \left ( f_i + \bfw^\top \bfz_i^\Psi \right )$. The Lagrange function of this constrained problem is,

\begin{equation}
\label{equ:Lagrange}
\begin{aligned}
\mathcal{L}& = \frac{1}{n} \sum_{i=1}^n \xi_i + \frac{C_1}{2} \|\bfw\|_2^2  +  \frac{C_2}{2} \sum_{k=1}^\kappa \|\bfpsi_k\|_2^2 \\
& -\sum_{i=1}^n\alpha_i\xi_i - \sum_{i=1}^n\beta_i \left ( \xi_i - 1 + y_i \left ( f_i + \bfw^\top \bfz_i^\Psi \right ) \right ).
\end{aligned}
\end{equation}
Thus the dual form of the problem is given as

\begin{equation}
\label{equ:Lagrange1}
\begin{aligned}
\max_{\alpha_i|_{i=1}^n,\beta_i|_{i=1}^n} ~& \min_{\Psi,\bfw,\xi_{i=1}^n}  \left \{
\mathcal{L} = \frac{1}{n} \sum_{i=1}^n \xi_i + \frac{C_1}{2} \|\bfw\|_2^2  +  \frac{C_2}{2} \sum_{k=1}^\kappa \|\bfpsi_k\|_2^2 \right.  \\
& \left . -\sum_{i=1}^n\alpha_i\xi_i - \sum_{i=1}^n\beta_i \left ( \xi_i - 1 + y_i \left ( f_i + \bfw^\top \bfz_i^\Psi \right ) \right ) \right \}\\
s.t.~& \alpha_i \geq 0, \beta_i \geq 0, \forall ~i=1,\cdots,n.
\end{aligned}
\end{equation}
To solve this problem, we set the partial derivatives of $\mathcal{L}$ with regard to $\bfw$, $\xi_i$ to zeros,

\begin{equation}
\label{equ:Lagrange2}
\begin{aligned}
&\frac{\partial \mathcal{L}}{\partial \bfw}  = C_1 \bfw - \sum_{i=1}^n\beta_i   y_i  \bfz_i^\Psi = 0,\\
&\Rightarrow \bfw = \frac{1}{C_1} \sum_{i=1}^n\beta_i   y_i \bfz_i^\Psi,\\
&\frac{\partial \mathcal{L}}{\partial \xi_i}  = \frac{1}{n} -\alpha_i - \beta_i= 0,\\
&\Rightarrow \frac{1}{n} - \beta_i= \alpha_i \geq 0,\\
&\Rightarrow \frac{1}{n} \geq \beta_i.\\
\end{aligned}
\end{equation}
We substitute the results of (\ref{equ:Lagrange2}) to (\ref{equ:Lagrange1}), and obtain the following problem,

\begin{equation}
\label{equ:Lagrange3}
\begin{aligned}
\max_{\beta_i|_{i=1}^n} ~& \min_{\Psi}  \left \{
\mathcal{L} =  \frac{C_1}{2} \|\bfw\|_2^2  +  \frac{C_2}{2} \sum_{k=1}^\kappa \|\bfpsi_k\|_2^2 \right.  \\
& \left .  +\sum_{i=1}^n\beta_i \left ( 1 - y_i  f_i\right )
- \sum_{i=1}^n\beta_i  y_i  \bfw^\top \bfz_i^\Psi
 \right .\\
&= \left.  \frac{C_2}{2} \sum_{k=1}^\kappa \|\bfpsi_k\|_2^2  +  \sum_{i=1}^n\beta_i \left ( 1 - y_i  f_i\right )
- \frac{1}{2C_1}\sum_{i,j=1}^n\beta_i \beta_j y_i y_j  {\bfz_i^\Psi}^\top \bfz_j^\Psi
 \right \}\\
s.t.~& \frac{1}{n} \geq \beta_i \geq 0, \forall ~i=1,\cdots,n.
\end{aligned}
\end{equation}
To solve this problem, we use an iterative algorithm with alternate optimization strategy. We only consider two variables in (\ref{equ:Lagrange3}), which are $\beta_i|_{i=1}^n$ and $\Psi$, and all other variables have vanished. To optimize them, in an iteration, we first fix $\Psi$ and update $\beta_i|_{i=1}^n$, then with the updated $\beta_i|_{i=1}^n$, we update $\Psi$.

\subsubsection{Updating $\beta_i|_{i=1}^n$}

When $\Psi$ is fixed, the problem in (\ref{equ:Lagrange3}) is reduced to

\begin{equation}
\label{equ:beta}
\begin{aligned}
\max_{\beta_i|_{i=1}^n} ~&
\left\{  \frac{C_2}{2} \sum_{k=1}^\kappa \|\bfpsi_k\|_2^2  +  \sum_{i=1}^n\beta_i \left ( 1 - y_i  f_i\right )
- \frac{1}{2C_1}\sum_{i,j=1}^n\beta_i \beta_j y_i y_j  {\bfz_i^\Psi}^\top \bfz_j^\Psi
 \right \}\\
s.t.~& \frac{1}{n} \geq \beta_i \geq 0, \forall ~i=1,\cdots,n.
\end{aligned}
\end{equation}
This is a maximization problem with regard to $\beta_i|_{i=1}^n$. The objective is a quadratic function and the constraints are linear functions. Thus it is a linear constrained quadratic programming problem. We can solve it by active set algorithm.

\subsubsection{Updating $\Psi$}

When $\beta_i|_{i=1}^n$ is fixed, the problem in (\ref{equ:Lagrange3}) is reduced to

\begin{equation}
\label{equ:Psi1}
\begin{aligned}
\min_{\Psi} ~& \left \{  \frac{C_2}{2} \sum_{k=1}^\kappa \|\bfpsi_k\|_2^2  +  \sum_{i=1}^n\beta_i \left ( 1 - y_i  f_i\right )
- \frac{1}{2C_1}\sum_{i,j=1}^n\beta_i \beta_j y_i y_j  {\bfz_i^\Psi}^\top \bfz_j^\Psi
\right \}.
\end{aligned}
\end{equation}
We further rewrite the bag-level feature vector as

\begin{equation}
\begin{aligned}
\bfz_i^\Psi &= \left[\max_{j=1}^{m_i} \bfpsi_1^\top\bfx^i_j, \cdots, \max_{j=1}^{m_i} \bfpsi_\kappa^\top\bfx^i_j\right]^\top \\
&= \left[\bfpsi_1^\top\bfx^i_{\pi^i_1}, \cdots,\bfpsi_\kappa^\top\bfx^i_{\pi^i_\kappa}\right]^\top,
\end{aligned}
\end{equation}
where

\begin{equation}
\label{equ:pi}
\begin{aligned}
\pi^i_k = {\arg\max}_{j=1}^{m_i} \bfpsi_k^\top\bfx^i_j
\end{aligned}
\end{equation}
is the index of the instance which gives the maximum product between $\bfpsi_k$ and the instance. Thus we can rewrite ${\bfz_i^\Psi}^\top \bfz_j^\Psi$ in (\ref{equ:Lagrange3}) as follows,

\begin{equation}
\label{equ:zz}
\begin{aligned}
{\bfz_i^\Psi}^\top \bfz_j^\Psi
&= \sum_{k=1}^\kappa \left ( \bfpsi_k^\top\bfx^i_{\pi^i_k} \right ) \left ( \bfpsi_k^\top\bfx^j_{\pi^j_k} \right )\\
&= \sum_{k=1}^\kappa  \bfpsi_k^\top \left ( \bfx^i_{\pi^i_k} {\bfx^j_{\pi^j_k}}^\top \right ) \bfpsi_k .
\end{aligned}
\end{equation}
Substituting (\ref{equ:zz}) to (\ref{equ:Psi1}), we can rewrite it as follows,

\begin{equation}
\label{equ:Psi2}
\begin{aligned}
\min_{\Psi} ~& \left \{  \frac{C_2}{2} \sum_{k=1}^\kappa \|\bfpsi_k\|_2^2  +  \sum_{i=1}^n\beta_i \left ( 1 - y_i  f_i\right )\right .\\
&- \frac{1}{2C_1}\sum_{i,j=1}^n\beta_i \beta_j y_i y_j  \left [ \sum_{k=1}^\kappa  \bfpsi_k^\top \left ( \bfx^i_{\pi^i_k} {\bfx^j_{\pi^j_k}}^\top \right ) \bfpsi_k  \right ]\\
&=\sum_{k=1}^\kappa  \left [ \frac{C_2}{2} \|\bfpsi_k\|_2^2
- \frac{1}{2C_1} \bfpsi_k^\top \left ( \sum_{i,j=1}^n\beta_i \beta_j y_i y_j \bfx^i_{\pi^i_k} {\bfx^j_{\pi^j_k}}^\top \right ) \bfpsi_k  \right ] \\
& \left .+  \sum_{i=1}^n\beta_i \left ( 1 - y_i  f_i\right )\right \}.
\end{aligned}
\end{equation}
It is apparently that the objective function is composed of a submission of sub-functions over individual codewords and a term irrelevant to the codewords. Thus we can optimize each codewords one by one independently. We define the objective for each $\bfpsi_k$ as follows,

\begin{equation}
\label{equ:Psi3}
\begin{aligned}
h(\bfpsi_k)= \frac{C_2}{2} \|\bfpsi_k\|_2^2
- \frac{1}{2C_1} \bfpsi_k^\top \left ( \sum_{i,j=1}^n\beta_i \beta_j y_i y_j \bfx^i_{\pi^i_k} {\bfx^j_{\pi^j_k}}^\top \right ) \bfpsi_k,
\end{aligned}
\end{equation}
which is a quadratic function of $\bfpsi_k$. The problem of (\ref{equ:Psi2}) can be rewritten as follows,

\begin{equation}
\label{equ:Psi4}
\begin{aligned}
\min_{\Psi} ~& \left \{
\sum_{k=1}^\kappa h(\bfpsi_k) + \sum_{i=1}^n\beta_i \left ( 1 - y_i  f_i\right )\right \}.
\end{aligned}
\end{equation}
Ignoring the last term irrelevant to the dictionary learning, this problem can be decomposed to $\kappa$ independent sub-problems,

\begin{equation}
\label{equ:Psi5}
\begin{aligned}
&\min_{\bfpsi_k} h(\bfpsi_k),k=1,\cdots,\kappa.
\end{aligned}
\end{equation}
To solve each of these problems, we use the sub-gradient deselect algorithm. Firstly, we update the instance indexes according to (\ref{equ:pi}) by using previous updated codewords, and then fix them to update the codewords themselves. To update them, we fist calculate the sub-gradient function of $h(\bfpsi_k)$ with regard to $\bfpsi_k$,

\begin{equation}
\label{equ:h}
\begin{aligned}
\nabla h(\bfpsi_k) = C_2\bfpsi_k
- \frac{1}{C_1} \left ( \sum_{i,j=1}^n\beta_i \beta_j y_i y_j \bfx^i_{\pi^i_k} {\bfx^j_{\pi^j_k}}^\top \right ) \bfpsi_k.
\end{aligned}
\end{equation}
Then we descent each codeword to the sub-gradient direction,

\begin{equation}
\label{equ:h}
\begin{aligned}
\bfpsi_k \leftarrow \bfpsi_k - \eta \nabla h(\bfpsi_k) = \bfpsi_k - \eta \left ( C_2\bfpsi_k
- \frac{1}{C_1} \left ( \sum_{i,j=1}^n\beta_i \beta_j y_i y_j \bfx^i_{\pi^i_k} {\bfx^j_{\pi^j_k}}^\top \right ) \bfpsi_k \right ).
\end{aligned}
\end{equation}
where $\eta$ is the step size of the descent. The algorithm of updating the $k$-th codeword is summarized in Algorithm 1.

\begin{itemize}
\item Algorithm 1: \textbf{Updating the $k$-th codeword}.
\begin{enumerate}
\item $\bfpsi_k \leftarrow [0,\cdots,0]^\top$;
\item \textbf{For} $t=1,\cdots,T$
\item ~~~~~$\pi^i_k = {\arg\max}_{j=1}^{m_i} \bfpsi_k^\top\bfx^i_j$ for $i=1,\cdots,n$;
\item ~~~~~$\bfpsi_k \leftarrow \bfpsi_k - \eta \left ( C_2\bfpsi_k
- \frac{1}{C_1} \left ( \sum_{i,j=1}^n\beta_i \beta_j y_i y_j \bfx^i_{\pi^i_k} {\bfx^j_{\pi^j_k}}^\top \right ) \bfpsi_k \right )$;
\item \textbf{End of For}
\item Output $\bfpsi_k$.
\end{enumerate}
\end{itemize}

\subsection{Iterative algorithm}
\label{sec:algorithm}

The overall algorithm for the learning of domain transfer dictionary and its corresponding adaptation function is summarized in Algorithm 2. In this  algorithm, the domain transfer dictionary and the Lagrange multiplier variables are updated alternately in a while loop. After the while loop is completed, the adaptation function parameter is recovered from the Lagrange multiplier variables.

\begin{itemize}
\item Algorithm 2: \textbf{Learning domain transfer dictionary and adaptation function parameter (DTC)}.
\begin{enumerate}
\item \textbf{Input}: target training set, $\{(B_i,y_i)\}_{i=1}^n$;
\item \textbf{Input}: source training dictionary and its corresponding classifier parameter, $\Phi$ and $\bfv$;
\item \textbf{Input}: size of the domain transfer dictionary, $\kappa$;
\item \textbf{Input}: weights of regularization terms, $C_1$ and $C_2$.

\item \textbf{Initialization}: $f_i = \bfv^\top \bfz_i^\Phi$ for $i=1,\cdots,n$.
\item \textbf{Initialization}: $t=1$;
\item \textbf{Initialization}: Initialize the domain transfer dictionary $\Psi^1=\{\bfpsi_k^1\}|_{k=1}^\kappa$.

\item \textbf{Repeat}

\begin{enumerate}

\item \textbf{For $i=1,\cdots,n$}
\item ~~~~Update ${\bfz_i^{\Psi^t}} = \left[\max_{j=1}^{m_i} {\bfpsi_1^t}^\top\bfx^i_j, \cdots, \max_{j=1}^{m_i} {\bfpsi_\kappa^t}^\top\bfx^i_j\right]^\top$;
\item \textbf{End of For}

\item Update $\beta_i^t|_{i=1}^n$ by solve the quadratic programming problem by fixing ${\bfz_i^{\Psi^t}}|_{i=1}^n$;

\begin{equation}
\begin{aligned}\beta_i^t|_{i=1}^n =
{\arg\max}_{\beta_i|_{i=1}^n} ~&
\left\{ \sum_{i=1}^n\beta_i \left ( 1 - y_i  f_i\right )\right.\\
&\left.
- \frac{1}{2C_1}\sum_{i,j=1}^n\beta_i \beta_j y_i y_j  {{\bfz_i^{\Psi^t}}}^\top {\bfz_j^{\Psi^t}}
 \right \},\\
s.t.~& \frac{1}{n} \geq \beta_i \geq 0, \forall ~i=1,\cdots,n.
\end{aligned}
\end{equation}

\item \textbf{For $k=1,\cdots,\kappa$}
\item ~~~~Update $\bfpsi_k^{t+1}$ by fixing $\beta_i^t|_{i=1}^n$ and using Algorithm 1;
\item \textbf{End of For}

\item $t=t+1$.

\end{enumerate}

\item \textbf{Until convergency}

\item Update $\bfw = \frac{1}{C_1} \sum_{i=1}^n\beta_i^{t+1}   y_i \bfz_i^{\Psi^{t+2}},$;

\item \textbf{Output}: $\Psi^{t+2}$, $\bfw$.

\end{enumerate}
\end{itemize}

\section{Experimental results}
\label{sec:experiment}

In this section, we use three benchmark data sets to evaluate the performance of the proposed algorithm. The method is compared to state-of-the-arts domain adaptation algorithms, and also some single domain dictionary learning algorithms. The running time of the algorithm is reported, and the convergency of the iterative algorithm is also plot.

\subsection{Benchmark data sets}

We use two benchmark domain transfer multi-instance data sets in our experiments, which are introduced as follows.

\subsubsection{TRECVID data set}

The first data set used is the TRECVID data set \cite{Oomen201391,Awad2014}. This data set is a set of key frames of video programs. The total number of the key frames is 61,901, and they belongs to 36 classes of concepts. This data set is composed of two sub-sets, which are TRECVID 2005 data set and TRECVID 2007 data set. The differences of program structure and production values between these two sub-sets are significantly, thus we treat them as two different domains. Moreover, we choose the key frames of the Chinese channel CCTV4 from TRECVID 2005 as source domain data, and the entire TRECVID 2007 data set as the target domain \cite{duan2012domain}. To present each key frame, we extract the SIFT local features, and treat each key frame as a bag of the local features. Each local feature is treated as an instance. Thus this is a multi-instance learning problem.

\subsubsection{MRSC+VOC data set}

The seconde data is the combination of the MRSC data set and VOC. The MRSC data is an image data set containing 4,323 images of 18 classes, and the VOC data set is also an image data set containing 5,011 images of 20 classes. {The MRSC data set is publicly accessible at http://research.microsoft.com/enu\\
s/projects/objectclassrecognition, and the VOC data set is publicly accessible at http://pascallin.ecs.soton.ac.uk/challenges/VOC/voc2007. }
Both the two data sets share 6 common classes, which are listed as follows: aeroplane, bicycle, bird, car, cow, and sheep. However, the distractions of the data of these two data sets are significantly different, and we tree them as two different domains. Thus we can combine the images of these 6 classes of both the data sets to one cross-domain data set. The cross-domain data set is composed of a source domain set of MSRC containing 1,269 images, and a target domain of VOC containing 1,530 images \cite{long2013transfer}. Thus the total number of the images of this cross-domain data set is 2,799. To present each image, we also extract a set of SIFT local features, and treat each local feature as an instance. Each image is treated as a bag of instances.

\subsubsection{20 Newsgroups data set}

The third data set is a subset of the 20 Newsgroups data set. This data set is a set of 18,774 documents. The documents belong to 6 main classes, and 20 sub-classes. To construct the setting of domain learning, we set the main class ``comp" to be the positive class, and the main class ``rec" to be the negative class. In the positive class, we further select two sub-classes to be source domain and target domain respectively, i.e., ``comp.windows.x" and ``comp.sys.ibm.pc". For the negative class, we also select two sub-classes to be source domain and target domain, i.e., ``rec.sport.hockey" and ``rec.motorcycles". Thus the source domain contains the data of ``comp.windows.x" and ``rec.sport.hockey", and the target domain contains the data of ``comp.sys.ibm.pc" and ``rec.motorcycles". To represent each document, we treat each paragraph as a instance, and use the word-frequency feature as the feature of the instance. Thus each document is a bag of instances.

\subsection{Experimental protocol}

To conduct the experiment, we use the 10-fold cross-validation. The target domain data set is split to 10 sub-sets. Each sub-set is used as a training set, and the other 9 sub-sets are used as testing set. We fist train a source domain dictionary and classifier over the entire source domain data set, and then use the target training set to train the domain-transfer dictionary and its corresponding adaptive function parameter. Then the target domain predictor is obtained as the combination of the source domain predictor and the adaptive function. The target domain predictor is finally tested over the target domain test data set. Please note that only one fold is used as target domain training set. This is insufficient for training a good target domain predictor without the help of the source domain data. This setting makes it necessary to perform domain transfer learning. The average classification accuracy over the ten folds of training set is evaluated as performance measure.

\subsection{Experimental results}

In this section, we report the experimental results of the proposed algorithm.

\subsubsection{Comparison to state-of-the-arts}

We first compare the proposed algorithm, DTC, against both some common domain transfer learning methods and two domain transfer multi-instance learning methods. The compared domain transfer learning methods are listed as follows,

\begin{itemize}
\item Domain transfer SVM  (DT-SVM) \cite{duan2009domain},
\item Domain transfer multiple kernel learning (DT-MKL) \cite{duan2012domain},
\item Deep low-rank coding for transfer learning (DLRC) \cite{ding2015deep}, and
\item Cross-domain sparse coding (CDSC) \cite{wang2013cross}.
\end{itemize}
For these methods, the dictionary used to represent the bags are not learned cross domains. The dictionary is learned before the transfer learning is conducted and it is fixed during the transfer process. We also compare the proposed domain transfer multi-instance learning to two other such algorithms, which are

\begin{itemize}
\item the original multiple instance transfer learning (MITL) \cite{zhang2009multiple}, and
\item knowledge transfer learning for multiple instance learning (AKTL-MIL) \cite{wang2014adaptive}.
\end{itemize}
Please note that for the MITL algorithm, it assumes that there are several source domains, and the algorithm learns the predictors of different source domains and the combination coefficients of the predictors. However, in our experimental setting, we have only one source domain. Thus we only learn one source domain predictor and set its coefficient to one.

The comparisional results of different domain transfer learning algorithms are given in Table \ref{tab:transfer}. From Table \ref{tab:transfer} we observe that the proposed method outperforms all the other four compared domain transfer learning methods, including linear classifier \cite{duan2009domain}, multi-kernel classifier \cite{duan2012domain}, deep coding \cite{ding2015deep}, and sparse coding \cite{wang2013cross}. This indicates that for multi-instance domain transfer learning, using a dictionary is the most effective method to represent the multi-instance data. Moreover, the outperforming of the proposed method over the compared method is even more significant over the data sets of TRECVID and MRSC+VOC. For example, over TRECVID data set, DTC is the only algorithm that achieves an average accuracy higher than 0.300, and over MRSC+VOC, DTC is also the only algorithm that has a higher average accuracy higher than 0.400. Both these two data sets are image sets with local features. This implies that the proposed domain-transfer dictionary learning is especially suitable for computer vision tasks which represent images as collections of visual local features.

\begin{table}
\centering
\caption{Average classification accuracy of comparison among domain transfer learning methods}
\label{tab:transfer}
\begin{tabular}{|l||c|c|c|}
\hline
Method&TRECVID&MRSC+VOC&20 Newsgroups\\\hline\hline
DT-SVM&0.208&0.362&0.856\\\hline
DT-MKL&0.268&0.397&0.918\\\hline
DLRC&0.196&0.384&0.862\\\hline
CDSC&0.209&0.367&0.901\\\hline
DTC&\textbf{0.301}&\textbf{0.435}&\textbf{0.932}\\\hline
\end{tabular}
\end{table}

The results of different domain transfer multi-instance learning methods are given in Table \ref{tab:multi}. From Table \ref{tab:multi}, we observe that the proposed DTC algorithm also outperforms the two existing multi-instance transfer learning algorithms, especially when compared to MITL. This is not surprising, because MITL conduct the adaptation to the target domain by weighting the predictors of different source domains. However, in our experimental setting, there is only one domain, thus the domain adaptation is not accessible. AKTL-MIL also use bag-instance similarity to represent the bags, however, the dictionary is not learned. Our algorithm learns the adaptive function and the dictionary by transferring from the source domain to the target domain, and thus it is natural to obtain a good result.

\begin{table}
\centering
\caption{Average classification accuracy of comparison among domain transfer multi-instance learning methods}
\label{tab:multi}
\begin{tabular}{|l||c|c|c|}
\hline
Method&TRECVID&MRSC+VOC&20 Newsgroups\\\hline\hline
MITL&0.260&0.343&0.837\\\hline
AKTL-MIL&0.293&0.399&0.901\\\hline
DTC&\textbf{0.301}&\textbf{0.435}&\textbf{0.932}\\\hline
\end{tabular}
\end{table}

\subsubsection{Running time}

We are also interested in the running time of the compared algorithms. The running time of the compared domain transfer learning methods are reported in Table \ref{tab:Timetransfer}. From Table \ref{tab:Timetransfer}, we observe that the proposed DTC method is the most time-consuming method, but it is still comparable to the other domain transfer learning methods. The reason is its additional process of learning the dictionary beside the domain transfer classifier. Actually, the other methods take the bag-level features as input, and thus the time of constructing the dictionary is ignored. Thus it is not a fair comparison. Moreover, considering the significant improvement of classification accuracy archived by DTC, it is still the best option over the domain transfer multi-instance learning problem. The least time-consuming method is a sparse coding method, CDSC. It also learns a dictionary, but this dictionary is used to represent the bag-level features, and it takes the bag-level features as inputs. Thus the multi-instance learning process is not involved in the learning process. Moreover, over the three data sets, the running time for the TRECVID data set is the longest, due to its large size.

\begin{table}
\centering
\caption{Running time (second) of comparison among domain transfer learning methods}
\label{tab:Timetransfer}
\begin{tabular}{|l||c|c|c|}
\hline
Method&TRECVID&MRSC+VOC&20 Newsgroups\\\hline\hline
DT-SVM&1682&654&482\\\hline
DT-MKL&1812&722&587\\\hline
DLRC&947&365&148\\\hline
CDSC&861&247&130\\\hline
DTC&2043&510&254\\\hline
\end{tabular}
\end{table}

We also compare the proposed method against the two domain transfer multi-instance learning method regarding the running time performance. The compared methods include MITL and AKTL-MIL. The results are reported in Table \ref{tab:Timemulti}. It seems that MITL is comparable with DTC in the respect of running time, but AKTL-MIL is more time-consuming than DTC. This is not surprising because AKTL-MIL is a more complex algorithm than DTC.

\begin{table}
\centering
\caption{Running of comparison among domain transfer multi-instance learning methods}
\label{tab:Timemulti}
\begin{tabular}{|l||c|c|c|}
\hline
Method&TRECVID&MRSC+VOC&20 Newsgroups\\\hline\hline
MITL&1923&521&198\\\hline
AKTL-MIL&2684&721&364\\\hline
DTC&2043&510&254\\\hline
\end{tabular}
\end{table}

\subsection{{Analysis of sensitivity of the algorithm to $C_1$ and $C_2$}}

\begin{figure}
\centering
\includegraphics[width=0.65\textwidth]{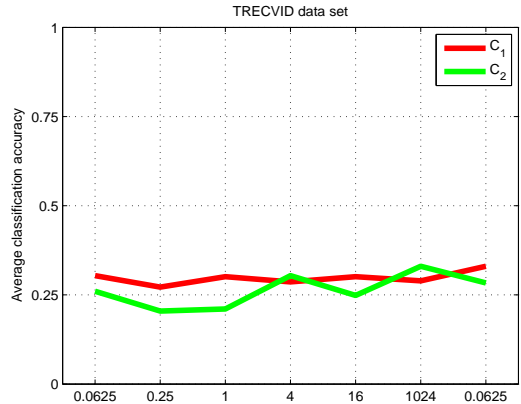}\\
\caption{Curves of changes of accuracy to changes of $C_1$ and $C_2$.}
\label{fig:C1C2}
\end{figure}

{We perform the analysis of the sensitivity of the proposed algorithm to the two tradeoff parameters, $C_1$, and $C_2$. We plot the average classification accuracy of the proposed algorithm with different values of $C_1$ and $C_2$ over the TRECVID data set in Fig. \ref{fig:C1C2}. According to the figure, the proposed algorithm are stable to the changes of both the two parameters.}

\section{Conclusions}
\label{sec:conclusion}

In this paper, we study the problem of domain transfer learning with multi-instance learning. We proposed to use the multi-instance dictionary to represent the bag of instances. However, due to the significant difference between the distributions of source and target domains, the dictionary learned from the source domain is not suitable to the target domain. We propose to learn a domain transfer dictionary to solve this problem. A target domain bag is represented by both the source domain and the domain transfer dictionaries, and then classified by the source domain classifier and an adaptive function simultaneously. The target domain classification response is the combination of the source domain response and the result of the domain adaptive function. To learn the domain transfer dictionary and its corresponding adaptive function parameter, we model a minimization problem, which minimizes the complexities of the adaptive function parameters, the domain transfer dictionary, and the average classification error jointly. The optimization problem is solved by an iterative algorithm. Our proposed method is shown to be effective by experiments over three benchmark data sets. In the future, we will also use the proposed algorithm to applications of bioinformatics \cite{peng2015modeling,zhou2014biomarker,xu2016mechanical,wang2014computational,liu2013structure} and data analysis of nanotechnology \cite{liu2016optofluidic,liu2015electro,liu2014correlated,liu2013effect,chen2011dual}.


\end{document}